\pdfoutput=1

\documentclass[11pt]{article}

\usepackage[]{ACL2023}

\usepackage{times}
\usepackage{latexsym}
\usepackage{amsmath} 
\usepackage{booktabs}
\usepackage{url}
\usepackage{multirow}
\usepackage{booktabs}
\usepackage{enumitem}
\usepackage{nicematrix}
\usepackage{CJKutf8}
\usepackage{longtable}
\usepackage{tabularx}
\usepackage{graphics}
\usepackage{graphicx}
\usepackage{colortbl}
\usepackage{xcolor, soul}
\usepackage{color}
\usepackage{transparent}
\usepackage{makecell}
\usepackage{amsmath}
\usepackage{ulem}
\usepackage{arydshln}
\newcommand{\comment}[1]{}
\usepackage{CJKutf8}

\usepackage[T1]{fontenc}

\usepackage[utf8]{inputenc}

\usepackage{microtype}

\usepackage{inconsolata}

\usepackage{dblfloatfix}

\title{Comparing Hallucination Detection Methods for Multilingual Generation}

\author{Haoqiang Kang \quad Terra Blevins \quad Luke Zettlemoyer \\
        Paul G. Allen School of Computer Science \& Engineering,\\ University of Washington \\
        {\tt \{haoqik, blvns, lsz\}@cs.washington.edu}}

\begin{document}
\maketitle
\begin{abstract}
While many hallucination detection techniques have been evaluated on English text, their effectiveness in multilingual contexts remains unknown. This paper assesses how well various factual hallucination detection metrics (lexical metrics like ROUGE and Named Entity Overlap, and Natural Language Inference (NLI)-based metrics) identify hallucinations in generated biographical summaries across languages. We compare how well automatic metrics correlate to each other and whether they agree with human judgments of factuality. Our analysis reveals that while the lexical metrics are ineffective, NLI-based metrics perform well, correlating with human annotations in many settings and often outperforming supervised models. However, NLI metrics are still limited, as they do not detect single-fact hallucinations well and fail for lower-resource languages. Therefore, our findings highlight the gaps in exisiting hallucination detection methods for non-English languages and motivate future research to develop more robust multilingual detection methods for LLM hallucinations.\footnote{The code and annotated dataset will be released upon publication.}
\end{abstract}

\section{Introduction}

Large Language Models (LLMs) have made remarkable advances in text generation. However, they are still prone to hallucinating facts, or generating text that conflicts with established world knowledge~\cite{huang2023survey, zhang2023siren}. While there has been considerable research towards detecting hallucinations in English~\cite{huang2023survey, zhang2023siren, ji2023survey}, much less focus has been given to multilingual hallucinations. Therefore, it is currently unclear whether the methods developed for detecting and addressing hallucinations in English are effective or even applicable in multilingual settings.

This paper evaluates the effectiveness of various automatic metrics, initially proposed for English factual hallucination detection, within a multilingual context. We focus on \textit{automatic} metrics requiring minimal in-language resources to perform hallucination detection, as this makes them most readily applicable to new languages; these metrics include traditional lexical metrics, such as ROUGE~\cite{lin2004rouge} and Named Entity Overlap, as well as Natural Language Inference (NLI) metrics. We also consider the differences between reference-based metrics and pairwise metrics based on the consistency among generated samples. To evaluate these metrics, we present correlation studies comparing these automated metrics directly, against supervised hallucination detection methods, and with human judgments of generation factuality.

We empirically evaluate these hallucination detection techniques in the multilingual context with a new dataset of parallel biographical generations (Section \ref{sec:bio-gen}). Our experiments find that:
    (1) lexical overlap metrics do not agree with NLI metrics \textit{or} human judgments when detecting hallucinations in reference or pairwise settings;
    (2) while pairwise NLI metrics strongly correlate with reference-based ones in high-resource languages, this significantly diminishes in low-resource settings;
    (3) automatic NLI metrics effectively detect sentence-level hallucinations in high-resource languages when compared to human evaluations, but not when assessing atomic facts; and
    (4) NLI metrics outperform supervised approaches at detecting hallucinations that can be verified or refuted by the reference text, but not on unverifiable errors.

Overall, while lexical overlap methods and pairwise comparisons of generated texts are more accessible for evaluating low-resource languages, 
they are often inadequate at hallucination detection. Additionally, while NLI-based metrics can detect factual hallucinations --- and even outperform models trained on hallucination detection in some cases --- these metrics perform best on high-resource languages. This highlights that multilingual hallucination detection performance is closely tied to the availability and quality of language resources, mirroring the trend observed in English that detection depends on natural language understanding abilities~\cite{manakul2023selfcheckgpt, min2023factscore}. This points to a substantial gap in hallucination detection in multilingual and low-resource contexts and the need for future work bridging this divide.

\section{Multilingual Hallucination Detection}

We measure the efficacy of different automatic metrics on detecting multilingual hallucinations. We focus on biography generation, a domain that is particularly sensitive to factual accuracy and coherence \cite{min2023factscore, dhuliawala2023chain}. We test a suite of automatic metrics, each of which caters to a different aspect of factual generation: ROUGE \cite{lin2004rouge}, named entity overlap, and Natural Language Inference (NLI)-based methods.

\subsection{Multilingual Biography Generation}
\label{sec:bio-gen}

Inspired by prior work measuring factuality in English \cite{min2023factscore}, we generate parallel biographies in different languages. The generated texts are then compared against a reference text (for \textit{reference-based} metrics) and other generated samples (\textit{pairwise} metrics) to detect hallucinations.

This section characterizes the generation quality of these biographies (Table \ref{tab:lang_acc}). We consider the average length of each biography (in tokens and sentences), along with estimates of how accurate the generation language is to the prompt language, as in some cases, multilingual LMs will generate continuations in an unexpected language \cite{kang2023translate, bawden2023investigating}.

The length of the generated texts varies notably across languages. While high-resource languages like English and French generate longer outputs, mid-resource languages such as Thai tend to generate much shorter biographies and incomplete sentences. Low-resource languages fare even worse (for instance, Ukrainian averages just 5.7 tokens and 0.40 sentences), demonstrating the significant gap in generation abilities across languages.

We assess the accuracy of the generated languages through three metrics: the percentage of valid generations that is detectable for the \textit{langdetect} package (Valid \%), 
the most frequently generated language for a given target language (Flang), and the accuracy of generated language out of the valid generations (Acc.). For high-resource languages like English, Chinese, Spanish, and French, the models generally generate text in the correct language; however, for the languages highlighted with an \uwave{underwave} the model generates in the wrong language the majority of the time. Often, this is due to the model generating in a closely related high-resource language. For languages such as Italian and Bulgarian, many inaccurate generations are in English. Similarly, Japanese generations often switch to Chinese when mistakes occur. Languages with more distinctive linguistic features ---such as Thai's unique script---facilitate more accurate model generations.


\begin{table}[ht]
    \centering
    \small
    \begin{tabular}{lccccc}
        \toprule
        Lang. & \#Token & \#Sent. & Valid \% & FLang. & Acc. \\
        \midrule
        \textbf{\underline{en}} & 78.3 & 2.64 & 99.97 & en & 96.0\\
        \textbf{\underline{zh}} & 115.8 & 4.30 & 100.00 & zh & 92.43  \\
        \textbf{\underline{es}} & 62.8 & 2.01 & 100.00 & es & 92.33  \\
        \textbf{\underline{fr}} & 71.3 & 2.24 & 100.00 & fr & 93.23  \\
        \textbf{\underline{vi}} & 45.6 & 1.66 & 98.92 & vi & 71.67 \\
        \textbf{\underline{id}} & 46.3 & 1.76 & 98.30 & \uwave{en} & 36.45 \\
        de & 63.3 & 2.33 & 99.58  & \uwave{en} & 2.79 \\
        it & 58.1 & 1.94 & 99.76  & \uwave{en} & 3.31 \\
        ja & 50.3 & 1.97 & 90.73 & \uwave{zh} & 21.85 \\
        bg & 17.4 & 1.15 & 86.74  & \uwave{en} & 13.69  \\
        ro & 9.6 & 0.93 & 80.24 & \uwave{en} & 2.68  \\
        \underline{sv} & 7.6 & 0.51 & 40.73 & \uwave{en} & 1.79 \\
        th & 14.8 & 0.81 & 77.08 & th & 94.96 \\
        ru & 10.2 & 0.68 & 55.49 & ru & 50.44\\
        uk & 5.7 & 0.40 & 35.24 & uk & 41.87 \\
        fa & 3.2 & 0.13 & 10.80 & ur & 29.90 \\
        fi & 1.7 & 0.11 & 9.76 & fi & 34.52 \\
        ko & 2.0 & 0.09 & 8.37  & ko & 47.30\\
        hu & 0.8 & 0.05 & 6.28 & pt & 14.36 \\
        \midrule
        Avg. & 34.8 & 1.35 & 31.65 & - & 50.01  \\
        \bottomrule
    \end{tabular}
    \caption{Quality statistics for BLOOMZ-mt generations. Languages that occur in the ROOTS pretraining corpus are in \textbf{bold} \cite{laurenccon2022bigscience}, and underlined languages are in the xP3mt fine-tuning dataset \cite{muennighoff-etal-2023-crosslingual}. "FLang." refers to the most frequently generated language for each prompt language.}
    \label{tab:lang_acc}
\end{table}

\subsection{Automatic Metrics}
\label{sec:automatic-metrics-def}
After quality verification of generated samples and filtering examples where the output is in an incorrect language, we compare the efficacy of different hallucination detection metrics on the remaining generations. We consider automatic metrics for detecting hallucinations in long-form generations that work by assessing the consistency between a target generation and either a reference text or its other generations. Specifically, we focus on metrics that \textit{do not} require supervised hallucination data: many languages do not have datasets available for this task, which makes these supervised methods infeasible for those settings.\footnote{However, for completeness we include two recent supervised methods for multilingual hallucination detection in \S \ref{sec:human-eval}.}

\paragraph{ROUGE}
The ROUGE metric is employed to assess the token-level similarity between texts. We consider the generated text's ROUGE 1 (R1) and L (RL) scores against the reference.

\paragraph{Named Entity Overlap (NEO)}
We hypothesize that the sets of named entities in the gold and generated text will differ if there is hallucination in the generation~\cite{nan-etal-2021-entity}. We calculate the F1, precision, and recall scores of named entities between the generated and reference text as an estimate for factual hallucinations.

\paragraph{NLI-based Detection}
Following \citet{manakul2023selfcheckgpt} and \citet{elaraby2023halo}, we adopt the NLI-based zero-shot sentence-level \textsc{SUMMAC} ($SummaC_{zs}$) scoring system \cite{laban2021summac} to evaluate hallucinations. The $SummaC_{zs}$ method was originally developed to gauge the consistency between a summary $S$ and a document $D$, by segmenting them into sentences $S_1, \dots, S_N$ and $D_1, \dots, D_M$ respectively. Aligning with the optimal configuration in \citet{laban2021summac}, we employ the $max$ operator to compute the score for a sentence. Denote $e_{S_n}^{D_m}$ and $c_{S_n}^{D_m}$ as the entailment and contradiction score for the generated sentence $S_n$ given the reference sentence $D_m$, respectively.

We define three metrics to quantify verifiable hallucination and one metric to quantify unverifiable hallucination, respectively. At sentence-level detection, for a generated sentence $S_i$ and a reference $D$, to detect verifiable hallucination, we define the following three metrics:
$\mathbf{ENT_{S_i}} = \max_{m}{e_{S_i}^{D_m}} $, 
$\mathbf{CON_{S_i}} = \max_{m}{c_{S_i}^{D_m}} $, and
$\mathbf{DIFF_{S_i}} = \max_{m}{e_{S_i}^{D_m}} - \max_{m}{c_{S_i}^{D_m}}$. 
To detect unverifiable hallucination, we define the following metric:
$$ \mathbf{UNV_{S_i}} = 1 - \max(\max_{m}{e_{S_i}^{D_m}}, \max_{m}{c_{S_i}^{D_m}}) $$

\noindent When evaluating each of the above hallucination metrics on a generated text $\hat{t}$, we consider two settings as the reference text $t$: \\
\noindent \textbf{Reference-based} This setting compares $\hat{t}$ against the relevant biographical article in Wikipedia. \\
\noindent \textbf{Pairwise} We generate $k$ samples for each biography. In this setting, we compare $\hat{t}$ against the other generated samples for the same person and calculate the average score across all generations. Experimental details for calculating these metrics are given in the Appendix.

\section{Experiment Setup}

\paragraph{Dataset}
Our curated dataset encompasses 19 languages: English, Spanish, Russian, Indonesian, Vietnamese, Persian, Ukrainian, Swedish, Thai, Japanese, German, Romanian, Hungarian, Bulgarian, French, Finnish, Korean, Italian, and Chinese. Using WikiData, we extract the names of 500 people who are covered by all of these languages on Wikipedia, based on diverse page view counts from 2022-01-01 to 2023-01-01. For our reference text, we use the Wikipedia API to obtain the full-page content. We detect instances where the LLMs generate text in an incorrect language with langdetect, which covers all 19 languages in our experiment.\footnote{APIs: \url{https://query.wikidata.org/}, \url{https://pypi.org/project/wikipedia/}, and \url{https://pypi.org/project/langdetect/}, respectively.}

\paragraph{Models and Prompting}

We generate text samples with the BLOOMZ-mt model, which is fine-tuned with machine-translated prompts \cite{workshop2023bloom}; at the time of our experiments, BLOOMZ-mt is the largest open-source, multilingual LM. We use nucleus decoding \cite{Holtzman2020The} with $top\_p = 0.9$, which is a common and realistic configuration used in other works in LLM hallucination~\cite{liu2023correction}, and generate five responses per prompt to evaluate the pairwise, intrinsic metrics. For each evaluation language, we generate a prompt template with Google Translate. The template in English is \textit{"Tell me a biography of <Name>."}; the templates translated into other languages are in Appendix (Figure~\ref{fig:biography_prompts}).

\begin{table*}
    \small
    \centering
    \begin{tabular}{lcccccccccccc}
    \toprule
    Language& R1-F1  & R1-P   & R1-R  & RL-F1  & RL-P   & RL-R  & N-F1 & N-P  & N-R & $\mathrm{DIFF}$ & $\mathrm{UNV}$ &  $\mathrm{ENT}$ \\ 
    \midrule
    \multicolumn{13}{c}{High-Resource Languages} \\
    English    & 1.83  & 87.58 & 0.94 & 1.40  & 72.41 & 0.72 & 4.27  & 53.41 & 2.26 & -0.60   & 0.19& 0.16 \\
    Chinese    & 6.43  & 57.34 & 3.76 & 5.59 & 51.73 & 3.26 & 4.69  & 35.27 & 2.79 & -0.62   & 0.21 & 0.16\\
    Spanish    & 2.77  & 85.86 & 1.47 & 2.19  & 72.39 & 1.16 & 3.28  & 48.48 & 1.76 & -0.51   & 0.18  & 0.28\\
    French     & 2.18  & 87.78 & 1.13 & 1.67 & 73.51 & 0.87 & 4.35  & 57.41 & 2.31 & -0.54   & 0.16 &  0.25\\
    Vietnamese & 6.82  & 92.92 & 4.22 & 5.34 & 85.87 & 3.21 & - & -       & -      & -0.49   & 0.15 & 0.34\\
    Indonesian & 7.51  & 68.51 & 4.87 & 5.44  & 55.28 & 3.52 & -  & -       & -      & -0.45   & 0.22 & 0.32 \\
    \midrule
    \multicolumn{13}{c}{Middle-Resource Languages} \\ 
    German     & 0.38  & 71.34 & 0.19 & 0.31  & 67.60 & 0.16 & 0.83  & 36.06 & 0.42 & -0.65   & 0.15 &  0.50\\
    Italian	&0.50	&69.13	&0.25	&0.42	&63.77	&0.21	&1.00	&30.26	&0.52&	-0.58 &	0.17 & 0.42\\
    Japanese   & 0.73  & 14.62 & 0.40 & 0.64  & 13.53  & 0.35 & 0.47  & 15.52 & 0.25 & -0.72   & 0.21 & 0.26\\
    Bulgarian  & 0.16  & 4.92  & 0.09 & 0.15  & 4.91  & 0.08 & -       & -       & -      & -0.61   & 0.19&0.50 \\
    Romanian   & 1.02  & 69.75 & 0.53 & 1.00  & 69.08 & 0.52 & 0.39  & 17.47 & 0.20 & -0.29   & 0.24 & 0.76\\
    Swedish    & 0.66  & 86.37 & 0.33 & 0.64  & 85.87 & 0.33 & 1.28  & 45.24 & 0.66 & -0.40   & 0.63& 0.79 \\
    \midrule
    \multicolumn{13}{c}{Low-Resource Languages} \\ 
    Thai       & 0.04  & 1.14  & 0.02 & 0.04  & 1.14  & 0.02 & -       & -       & -      & -0.56   & 0.38& 0.32  \\
    Russian    & 0.09  & 4.69  & 0.05 & 0.09  & 4.62  & 0.05 & 0.48  & 11.28 & 0.25 & -0.58   & 0.47 & 0.40 \\
    Ukrainian  & 0.04  & 1.53  & 0.02 & 0.03  & 1.52  & 0.02 & 0.70  & 20.64 & 0.36 & -0.53   & 0.66 & 0.51 \\
    Persian    & 0.00  & 0.00  & 0.00 & 0.00  & 0.00  & 0.00 & -       & -       & -      & -0.50   & 0.92&  0.38 \\
    Finnish    & 0.89  & 37.70 & 0.46 & 0.80  & 35.61& 0.41 & 0.58  & 23.71 & 0.30 & -0.59   & 0.91 & 0.33 \\
    Korean     & 0.18  & 6.58  & 0.09 & 0.18  & 6.57  & 0.09 & 0.24  & 8.48 & 0.12 &-0.53	&0.94&  0.25\\
    Hungarian  & 0.74  & 64.74 & 0.37 & 62.56 & 23.23 & 0.36 & -       & -       & -      & -0.53   & 0.97 & 0.51 \\
    \bottomrule
    \end{tabular}
    \caption{Results of different reference-based metrics for the BLOOMZ-mt model. "-" indicates the language is not covered by the Spacy NER tool. All of the ROUGE and Named Entity Overlap (N) results are in percentage (\%).}
    \label{tab:ref}
\end{table*}

\section{Multilingual Hallucination Metrics}
\label{sec:metric-analysis}
This section compares how different automatic metrics estimate hallucinations in our generated biographical corpus (Section \ref{sec:metric-eval}). We then perform a correlation study to test whether these metrics \textit{agree} when hallucination occurs (Section \ref{sec:metric-corr}). 

\subsection{Automatic Metrics}
\label{sec:metric-eval}

We first consider how different referenced-based automatic methods for detecting hallucination perform across languages on the generated biographical data from the BLOOMZ-mt model (Table \ref{tab:ref}).\footnote{We observe similar trends on pairwise metrics (Table \ref{tab:pairwise}).} We find that, unsurprisingly, these measures indicate increases in hallucination on middle- and low-resource languages (e.g., lower overlap with the reference, higher $\mathrm{UNV}$ scores). However, the NLI-based $\mathrm{DIFF}$ scores remain relatively stable regardless of language resourcefulness.

\begin{table}[ht]
    \centering \small
    \begin{tabular}{lccc}
    \toprule
    Language & \multicolumn{1}{c}{$\mathrm{ENT}$} &\multicolumn{1}{c}{$\mathrm{DIFF}$} & \multicolumn{1}{c}{$\mathrm{UNV}$} \\
    \midrule
    \multicolumn{4}{c}{\textit{\# examples in correct language > 1,000}}  \\
    \underline{English}     & 0.55 & 0.38 & 0.19 \\
    \underline{French}    & 0.52 & 0.40 &0.15 \\
    \underline{Chinese}     & \textbf{0.56} & \textbf{0.41} & 0.21  \\
    \underline{Spanish}     & 0.46 & 0.41 &0.17 \\
    \underline{Thai}        & 0.36 & 0.39 &\textbf{0.32} \\
    \underline{Vietnamese}  & 0.35 & 0.31 & \colorbox{gray}{0.00} \\
    Indonesian  & \colorbox{gray}{0.28} & 0.31 & 0.09 \\
    \midrule

    \multicolumn{4}{c}{\textit{\# examples in correct language < 1,000}}  \\
    \underline{Russian}     & 0.16 & 0.21 & 0.11 \\
    Japanese    & 0.37 & 0.40 &\colorbox{gray}{0.07} \\
    Ukrainian   & 0.23 & 0.19 & 0.17 \\
    \underline{Bulgarian}   & 0.42 & 0.32 &0.28 \\
    Korean      & \colorbox{gray}{0.05} &\colorbox{gray}{0.08} &\colorbox{gray}{-0.01} \\
    \midrule
    \multicolumn{4}{c}{\textit{\# examples in correct language < 100}}  \\
    Finnish     & \colorbox{gray}{0.09} & \colorbox{gray}{0.12} & \colorbox{gray}{0.01} \\
    Italian     & \colorbox{gray}{0.12} & \colorbox{gray}{0.13} & \colorbox{gray}{0.14} \\
    Persian     & \colorbox{gray}{0.13} & \colorbox{gray}{0.15} & \colorbox{gray}{0.02} \\
    \underline{German}      & 0.50 & 0.45 &\colorbox{gray}{0.11}\\
    Romanian    & \colorbox{gray}{0.00} & \colorbox{gray}{0.00} &\colorbox{gray}{0.14}  \\
    Hungarian   & \colorbox{gray}{0.24} & \colorbox{gray}{0.21} & \colorbox{gray}{0.11} \\
    Swedish     & \colorbox{gray}{0.30} & \colorbox{gray}{0.27} & \colorbox{gray}{-0.29} \\

    \bottomrule
    \end{tabular}
    \caption{The correlation between the reference-based NLI result and the pairwise NLI result across different languages. The languages with \underline{underline} are covered in the XNLI finetuing dataset. The numbers in \colorbox{gray}{gray} have the \textit{p-values} larger than 0.05. }
    \label{tab:pairwise_corr}
\end{table}

\paragraph{Lexical Overlap Metrics} We also note some specific trends within this metric type. For example, high-resource languages (English, Chinese, Spanish, French, Vietnamese, and Indonesian) exhibit particularly high recall scores, suggesting that the text generated in these languages has better coverage of the corresponding Wikipedia reference content. In contrast, lower-resource languages demonstrate significantly diminished recall.

Interestingly, languages where BLOOMZ-mt frequently produces incorrect language outputs (e.g., German and Italian) or empty or incomplete generations (e.g., Swedish and Hungarian) maintain relatively high precision scores in the higher-quality outputs we evaluate. While these generations seem to contain few explicit hallucinations, they often exclude many facts from the reference, as indicated by their correspondingly low recall scores. \label{precision}

\paragraph{NLI-based Metrics} All languages we consider obtain negative $\mathrm{DIFF}$ scores, including higher-resource languages like English and Chinese. This indicates a tendency towards contradictions in the generated text with their respective reference texts --- as measured by the NLI classifier. 

For the $\mathrm{UNV}$ scores, higher and middle-resource languages (ranging from English to Romanian in the table \ref{tab:ref}) fall within a similar range of 0.15 to 0.25. In contrast, low-resource languages that often produce empty or incomplete generations, such as Ukrainian, Persian, Finnish, and Korean, obtain much higher $\mathrm{UNV}$ scores. This implies that the $\mathrm{UNV}$ metric is sensitive to incomplete text generations and missing information and may indicate the model's generation errors beyond hallucination.

\subsection{Correlation Study Across Metrics}
\label{sec:metric-corr}
In this section, we conduct a correlation analysis to determine whether the considered metrics agree in measuring hallucination in multilingual contexts. This includes (1) the correlation between lexical hallucination metrics and NLI-based metrics, (2) the agreement of the four reference-based NLI metrics, and (3) the relationship between pairwise metrics and reference-based metrics.

\begin{figure}[ht]
    \centering
    \includegraphics[width=0.95\columnwidth]{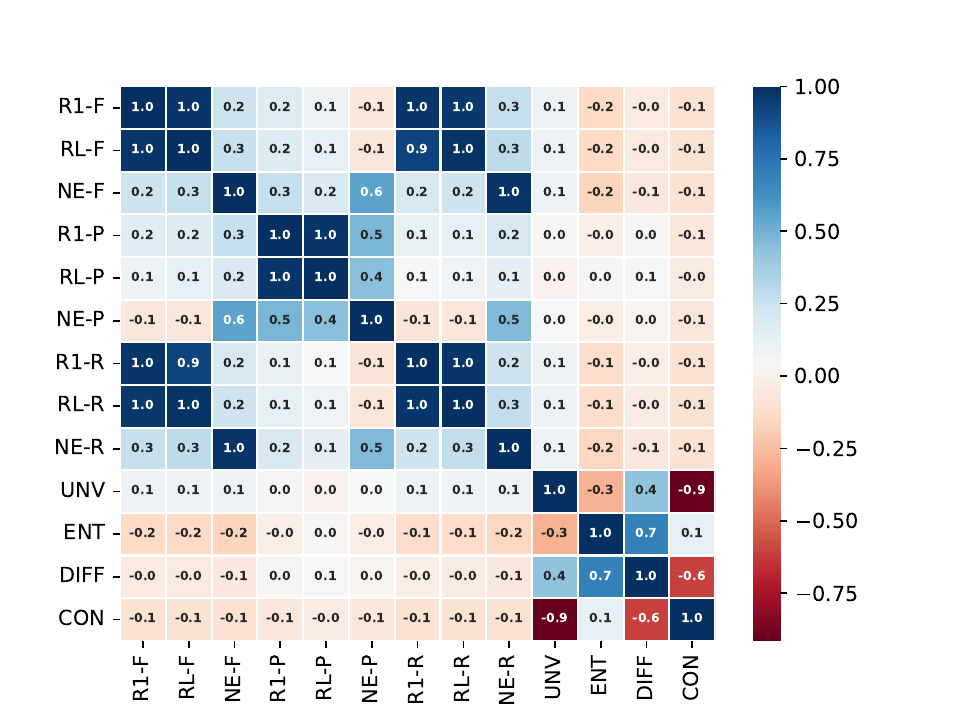}
    \caption{Heat map of the Pearson Correlation between reference-based metrics averaged over high-resource languages. All the P-values are less than 0.05.}
    \label{fig:heat_correlation}
    \vspace{-10pt} 
\end{figure}

\begin{table*}[ht]
    \centering
    \footnotesize
    \begin{tabular}{p{3cm}p{12cm}}
    \toprule
    \textbf{Entity} & \textbf{Generation} \\
    \midrule
    \multicolumn{2}{c}{\textit{Example of Annotation}}  \\
    Alessandro Del Piero & \colorbox{lightgray}{Gen:} Alessandro Del Piero, born on \colorbox{red}{September 28, 1976} in \colorbox{yellow}{Brescia, Italy}, is a \colorbox{green}{former Italian professional football player} who served as a \colorbox{green}{forward}. \colorbox{magenta}{Wiki:} Alessandro Del Piero, Italian male football player...The old Maldini, who was the head coach of the national team at the time, appointed newcomers Del Piero and Vieri as the main forwards... \colorbox{pink}{Comment:} There are 4 facts in this sentence, with 1 contradictory hallucination and 1 unverifiable hallucination. The birth date is wrong; Wikipedia doesn't mention the birth place; for the last entity, the evidence indirectly support it.\\
    
    \bottomrule
    \end{tabular}
    \caption{Example of human hallucination annotations. \colorbox{red}{Red} represents verifiable hallucinations contradicting evidence in the reference (\colorbox{magenta}{Wiki}), \colorbox{yellow}{yellow} denotes unverifiable hallucinations without relevant evidence, and \colorbox{green}{green} is supported by the reference text.}
    \label{tab:annot_example}
\end{table*}

\paragraph{Lexical hallucination metrics do not correlate with NLI-based metrics.} Figure~\ref{fig:heat_correlation} shows that in high-resource languages (i.e., English, Chinese, French, Spanish, Vietnamese, and Indonesian), ROUGE-1 and ROUGE-L metrics demonstrate a high degree of correlation, and Named Entity Overlap (NEO) correlates reasonably well with ROGUE precision metrics. However, we generally find no correlation between lexical- and NLI-based metrics, indicating that while both lexical- and NLI-based approaches are commonly proposed as automatic methods for hallucination detection, they do not measure the same deviations from a reference text. 

\paragraph{Reference-based NLI-based metrics.} We also observe interesting trends regarding the relationship between different NLI-based metrics (bottom right-hand corner of Figure~\ref{fig:heat_correlation}). We find that $\mathrm{ENT}$ scores are highly (inversely) correlated with the $\mathrm{DIFF}$ score, indicating that these metrics identify similar artifacts in the text. Moreover, we find a negative correlation between $\mathrm{UNV}$ and $\mathrm{CON}$ scores. This is because sentences that include verifiable hallucinations likely contradict the reference text. In contrast, sentences with information that is unsubstantiated by the reference (e.g., unverifiable) will be identified as neutral instead.

\paragraph{Pairwise and reference metrics do not correlate in low-resource languages.} For high-resource languages in the XNLI finetuning dataset (English, French, Chinese, Spanish, Bulgarian, and German), we observe higher correlations ranging between 0.35 to 0.56 for pairwise and reference-based NLI metrics when it comes to detecting verifiable hallucinations by $\mathrm{ENT}$ score (Table~\ref{tab:pairwise_corr}). This suggests pairwise metrics can identify generated content that deviates from the reference and may be useful for detecting hallucinations when gold reference texts are not available. However, the Pearson Correlation Coefficient shows lower correlation values (in the range of 0.15 to 0.21) when comparing pairwise and reference-based $\mathrm{UNV}$, indicating a less effective capture of extrinsic hallucinations involving plausible yet unverifiable information. 
For lower-resource languages, such as Finnish, Italian, Persian, correlation with the entailment score is often not statistically significant. This implies that the effectiveness of pairwise hallucination metrics is limited to higher-resource languages, highlighting the challenge of effective hallucination detection in limited resource contexts. 

\begin{table*}[!b]
    \centering
    \small
    \begin{tabular}{p{0.27\linewidth} p{0.68\linewidth}}
        \toprule
        \textbf{Question} & \textbf{Instructions} \\
        \midrule
        0. Atomic-Level Annotation & Extract a sentence that contains only one simple fact. \\
        1. Evidence Extraction & Copy and paste all relevant evidence. \\
        2. Total Facts & Provide an approximate count of the total facts. Each date is counted as one fact, except for birthdates and death dates, which are counted as two separate facts. \\
        3. Verifiable Correct Facts & Count the number of facts that can be verified as correct. \\
        4. Verifiable Contradictory Facts & Count the number of facts that contradict verified information. \\
        5. Unverifiable Facts & Count the number of facts that cannot be verified. \\
        6. Conflict with Preceding Context & Indicate whether there is any conflict with the preceding context (True or False). \\
        7. Conflict with Instructions & Determine whether there is any conflict with the instructions. Label it as False if the example provides (a) a biography of (b) the correct person. Otherwise, label it as True. \\
        \bottomrule
    \end{tabular}
    \caption{Instructions for manual hallucination annotations. Step 0 is only taken for atomic-level fact verification.}
    \label{tab:annotation_instructions}

\end{table*}

\section{Human Evaluation}
\label{sec:human-eval}
We manually annotate the model generations analyzed in the prior section; the annotations are performed on paired subsets of the English and Chinese generations by native speakers. Following the \textit{Attributable to Identified Sources (AIS)} paradigm~\cite{rashkin2023measuring} for measuring hallucination, annotators manually find all verifiable and unverifiable hallucinations by checking if the generated output is attributable to the Wikipedia reference at both the sentence- and atomic-fact-level. Table~\ref{tab:annot_example} shows example annotations.

\begin{table*}[ht]
\centering
\small

\begin{NiceTabular}{l|ccc|ccc}

\toprule
     & \multicolumn{3}{c}{Sentence Level} &  \multicolumn{3}{c}{Atomic-Fact Level} \\
    Metric  &  Pearson & AUC$_{\textit{F}}$ & AUC$_{\textit{NF}}$  & Pearson & AUC$_{\textit{F}}$ & AUC$_{\textit{NF}}$   \\ 
\midrule
Random      & -           & 10.84   & 82.86  & - &52.11&43.56\\ 
 \hdashline 
\textit{Pairwise} \\
R1-P. &  0.08$^\dagger$\       & 19.78   & 81.48  & 0.10  & 51.23 & 44.23   \\ 
RL-P. &  0.11$^\dagger$\        & 20.04   & 83.10 & 0.12$^\dagger$\ & 52.18 & 42.83 \\ 
NEO-P. &  0.14$^\dagger$\      & 17.49   & 80.84    & 0.09 & 53.09 & 45.32\\ 
$\mathrm{DIFF}$ & 0.21 & 38.46  & 89.49   & 0.19 &57.46 &54.41 \\ 
$\mathrm{ENT}$ & 0.31 & 40.32 & 90.86 & 0.23 & 60.71 & 57.48\\
$\mathrm{CON}$ & 0.11  & 16.47   & 80.41 & -0.01& 51.49 & 52.16 \\ 

\midrule
\textit{Reference} \\

R1-P. & 0.21        & 30.05   & 89.08   &  0.19 & 53.28 & 46.25  \\ 
RL-P. & 0.17        & 28.54   & 85.35    & 0.13$^\dagger$\ & 50.31 & 49.93  \\ 
NEO-P. & 0.17$^\dagger$\    & 16.15   & 83.75     & 0.12 & 57.54 & 47.51 \\ 
$\mathrm{DIFF}$    & 0.34        & 56.11   & 94.14 & 0.31 &65.85 &60.90  \\ 
$\mathrm{ENT}$ & \textbf{0.49} & \textbf{65.32} & \textbf{94.96}  & \textbf{0.35} &\textbf{68.00} & \textbf{63.69}  \\
$\mathrm{CON}$ & 0.08  & 31.56   & 87.49   &-0.19& 53.18 & 57.43  \\ 
$\mathrm{mFact}$ & 0.20  & 35.68   & 91.16   &0.29& 67.30& 61.67\\
$\mathrm{Seahorse}$ & -0.17$^\dagger$\  & 13.25   & 75.40   & -0.07$^\dagger$\ & 53.30 & 46.67 \\
\bottomrule
\end{NiceTabular}
\caption{Comparison of sentence and atomic-fact verifiable hallucination metrics with the human support rate. \textit{F} denotes factual examples and \textit{NF} denotes non-factual examples. $^\dagger$\textit{p-values} of this correlation is larger than 0.05.} 
\label{tab:veri-detection}
\vspace{-4pt}
\end{table*}

\subsection{Experimental Setup}

The authors (one per language) manually checked every sentence in the audited subset of generations, using the steps listed in Table \ref{tab:annotation_instructions}.
For atomic-fact-level annotation, a preprocessing step is taken to extract only sentences that contain a standalone proposition\footnote{A standalone proposition is independently interpretable from the information contained in the assertion.}~\cite{rashkin2023measuring}. 
Then, for both the sentence- and atomic-fact-level annotation, we annotate all relevant evidence sentences from the reference Wikipedia page and accumulate the counts for different types of propositions (Table~\ref{tab:annotation_instructions}). Table~\ref{tab:annot_stat} details data statistics from this annotation process.

\paragraph{Metrics} We compare our automatic metrics presented in \S \ref{sec:automatic-metrics-def} with human annotations using correlation and classification; we specifically compare precision metrics because they are generally the strongest automatic measure in Section~\ref{precision}. For correlation, we investigate the relationship of the metrics with the \textit{support rate} (SR; $N_{vs}/N_t$) for verifiable hallucination detection and with the \textit{unverified rate} ($N_{nv}/N_t$) for unverifiable hallucination detection using their \textul{Pearson} correlations.

To consider the classification agreement of these metrics, we calculate the Precision-Recall area under the curve (\textul{AUC-PR}) between the discretized human-annotated and automatic metrics. We convert human annotations into classification labels by labeling an example as factual only if all its facts are supported by evidence for verifiable hallucinations with the \textit{support rate} (Table \ref{tab:veri-detection}); for unverifiable hallucinations, we consider any sentence with at least one fact not supported or refuted by the reference to be \textit{unverified}: $N_{nv} \geq 1$ (Table~\ref{tab:unverify}). 

We discretize the automatic NLI-based metrics into classification labels by setting their respective thresholds, with 0.5 for the entailment and contradictory scores and 0 for the difference between these two scores. The thresholds were selected based on the different degrees of tolerance for the proportions of unverifiable hallucinations in a sentence. We then perform classification using the discretized human judgments as gold labels.


\subsection{Automatic Metric Results}

\paragraph{NLI entailment outperforms lexical metrics on sentence-level verification.} We observe low correlation between lexical metrics like ROUGE-1 (R1) and Named Entity Overlap (NEO) and the human-annotated support rate (SR; Table~\ref{tab:veri-detection}). Moreover, AUC-PR for these metrics exhibits minor improvements over the random classification baseline, particularly for non-factual examples (\textit{NF}), underscoring the limit of lexical metrics for accurately detecting factual hallucinations.

In contrast, the NLI-based metrics highly correlate with different measures of human verification for detecting verifiable hallucinations at the sentence level (Table~\ref{tab:veri-detection}). These results corroborate the NLI metric results in English in~\citet{manakul2023selfcheckgpt}. 
Specifically, the $\mathrm{ENT}$ and $\mathrm{DIFF}$ scores perform well, while $\mathrm{CON}$ alone does not demonstrate the same agreement with human judgments.


\paragraph{Pairwise metrics underperform reference-based metrics.} When compared against human annotations, pairwise metrics always underperform the reference-based ones: their Pearson correlations, AUC$_F$, AUC$_{NF}$, and accuracies are all worse than the metrics using references, and this holds at both the sentence- and atomic-fact-levels (Table~\ref{tab:veri-detection}). 
However, we note that pairwise NLI-based metrics, particularly the entailment and difference scores, demonstrate significantly better performance than the lexical-based pairwise metrics, though they still underperform comparable reference metrics. This suggests pairwise NLI-based approaches remain the most useful approach in hallucination detection settings where references are unavailable.

\paragraph{NLI metrics struggle to detect unverifiable hallucination and check atomic facts.} Comparison against human judgments on \textit{unverifiable} hallucination detection reveals that automatic metrics show only marginal improvements over random classification and low correlation with human annotations (Table \ref{tab:unverify}). 
Furthermore, NLI metrics also encounter significant challenges in accurately verifying the factuality of simple atomic facts (Table~\ref{tab:veri-detection}).
Both AUC-PR and accuracy demonstrate a marked decrease in the effectiveness of NLI-based metrics on atomic facts. This aligns with \citet{luo-etal-2022-simple-challenging}, which also highlighted similar limitations in English NLI metrics for verification. This finding underscores the need for alternative metrics to address current NLI approaches' limitations, especially in the critical area of atomic factuality evaluation.

\subsection{Supervised Metric Results}
We also consider two supervised metrics: \texttt{mFACT}~\cite{qiu2023detecting} and seahorse-Q4~\cite{clark2023seahorse}. \texttt{mFACT}~\cite{qiu2023detecting} automatically apply English hallucination detection approaches to new languages by first using faithfulness metrics to rank and annotate English examples, and then translating the most and least faithful samples into a set of target languages. They then train a classifier on each target language (including English and Chinese). The seahorse-Q4 metric~\cite{clark2023seahorse} similarly fine-tunes a mT5-large model~\cite{xue-etal-2021-mt5} on an attribution task, using attribution subset (Q4) of the SEAHORSE dataset.

\paragraph{Supervised metrics underperform NLI-based metrics in verifiable hallucination detection.} \texttt{mFACT} scores exhibit similar but slightly lower performance than NLI-based metrics at both the sentence and atomic fact levels (Table~\ref{tab:veri-detection}). 
The seahorse-Q4 approach shows comparable performance to \texttt{mFACT} in English (Appendix Table~\ref{tab:veri-detection-en}). 
However, as seahorse-Q4 does not see Chinese data in training, it fails to detect verifiable hallucinations in this setting, leading to a lower overall score and highlighting the brittleness of supervised methods for hallucination detection. 

\paragraph{Supervised metrics detect unverifiable hallucinations more effectively.} 
At the sentence level, \texttt{mFACT} has the highest correlation with the human unverified rate (Table~\ref{tab:unverify}). 
Furthermore, both Seahorse and \texttt{mFACT} exhibit better performance in AUC$_\text{V}$ than random prediction by 37\% and 10\% respectively -- outperforming the NLI-based metrics. The improvements over NLI-based methods here (but not on verifiable hallucinations) are surprising, given that these methods rely on reference-based supervision rather than intrinsic evaluation.

This trend also holds at the atomic-fact level, where the supervised metrics outperform other methods in across all measures. 
However, they still exhibit lower correlation with this unverified rate than at the sentence level, indicating that detecting hallucination on simpler atomic facts remains challenging in the unverified setting.




\begin{table}[t]
    \small
    \centering
    \begin{NiceTabular}{lccc}
        \toprule
        Metric  &  Pearson & AUC$_\textit{V}$ & AUC$_\textit{UNV}$ \\ 
        \midrule

        Random$_\textit{Sent}$ & -  &  36.54 & 60.63\\
        $\mathrm{UNV}$$_\textit{Sent}$ & 0.28 & 31.77 & \textbf{71.84} \\
        $\mathrm{mFACT}$$_\textit{Sent}$& \textbf{-0.29} & \textbf{54.17} &69.67 \\
        $\mathrm{seahorse}$$_\textit{Sent}$& \colorbox{gray!50}{-0.10} &42.08 &69.49 \\
        \hdashline
        Random$_\textit{Atom}$& - & 68.21 &24.90\\
        $\mathrm{UNV}$$_\textit{Atom}$& 0.12 &77.49 &27.36 \\
        $\mathrm{mFACT}$$_\textit{Atom}$& \textbf{-0.13} &\textbf{80.35} &\textbf{31.49}  \\
        $\mathrm{seahorse}$$_\textit{Atom}$& \colorbox{gray!50}{-0.09} &78.32 & 28.48  \\
        \bottomrule
    \end{NiceTabular}
    \caption{Agreement with human judgments on unverifiable hallucination detection, averaged for Chinese and English. $\textit{UNV}$ denotes unverifiable hallucination, and $\textit{V}$ denotes verifiable. \colorbox{gray!50}{Gray} values have \textit{p-values} $>$ 0.05.}
    \label{tab:unverify}
\end{table}

\section{Related Work}
\paragraph{Factuality Hallucination Detection}
Detecting hallucinations is crucial for ensuring the reliability of machine-generated content. One line of work leverages uncertainty by analyzing the probability of the LLM's output space \cite{mielke2022reducing, kadavath2022language, varshney2023stitch}; other methods evaluate consistency between repeatly generated samples (including NLI-based approaches)~\cite{elaraby2023halo, manakul2023selfcheckgpt}, similar to our pairwise evaluation setting. Similarly, our reference-based evaluations follow works that use external evidence as reference texts and verify if the generation is supported by this reference~\cite{chern2023factool, min2023factscore}. However, factuality hallucination evaluations are limited in multilingual settings, with existing works using prompting methods~\cite{ahuja2023mega} and machine translation~\cite{lai-etal-2023-chatgpt}. 


\paragraph{Task-specific Multilingual Hallucination}
The main focus of multilingual hallucination evaluation methods so far has been on specific downstream tasks. \texttt{mFACT} evaluates faithfulness in summaries by transferring English judgments into target languages via machine translation \cite{qiu2023detecting}. Similarly, \citet{aharoni2022mface} leverages factual consistency models to improve faithfulness in multilingual summarization. In neural machine translation (NMT), \citet{dale2022detecting} detect and alleviate hallucinations by measuring generation similarity to the source text; \citet{xu2023understanding} similarly study source effects on hallucination with input perturbations. Other works~\cite{lee2019hallucinations, raunak-etal-2021-curious} analyze the susceptibility of current NMT methods to generate hallucinations. 



\section{Conclusion}

This study investigates the effectiveness of automatic metrics for detecting factual hallucinations in non-English generations by considering these metrics directly (Section \ref{sec:metric-analysis}) and comparing their predictions to human judgments and supervised detection approaches (Section \ref{sec:human-eval}). We document that while traditional lexical metrics struggle to detect hallucinations in multilingual settings, NLI-based metrics show promise in high-resource languages at the sentence level. However, their effectiveness diminishes when applied to atomic facts, and the reliability of NLI-based metrics is tied to the performance of NLI models, posing a significant hurdle in lower-resource languages. Therefore, our analysis highlights that detecting hallucinations effectively in a language is directly linked to the availability and quality of linguistic resources in that language. As a result, automatically detecting hallucinations in lower-resource languages remains a significant challenge for current NLP methods. 

\section*{Limitations}
This study focuses on text generation and hallucination in a specific setting (namely, generating biographical summaries with the BLOOM-mt model) to perform a controlled study on how different automatic metrics detect factual hallucinations. It remains an open question whether these findings hold in other generation settings, particularly when there is less reliance on factual knowledge (such as story generation). 

Additionally, portions of our experimental setup rely on automatic methods. Specifically, we use machine translation to construct the prompt templates, which may introduce noise. Furthermore, due to the unavailability of native speakers for other languages, our human evaluation and comparison against automated metrics is limited to Chinese and English. In the future, we would like to expand on these findings with human evaluations in other, lower-resourced languages to confirm how well the automatic detection methods hold up in these settings.



\bibliography{anthology,custom}
\bibliographystyle{acl_natbib}

\appendix

\section{Qualitative Analysis of Challenging Cases in Annotation}
\label{sec:appendix}

We identify four challenging categories of hallucination detection for annotators and NLI metrics (Table~\ref{tab:challenge}).
\begin{itemize}[leftmargin=*]
    \item \textit{Inferred}: Implicit fact connections between generation and evidence.
    \item  \textit{Subjective}: Generation contains subjective content, which is challenging for both human annotators and fact-based NLI models. 
    \item \textit{Nuanced Difference}: There are subtle distinctions between evidence and generated text, which is often missed by surface-level text classification in NLI models. 
    \item \textit{Temporal Information}: Generation contains time-sensitive information, which requires models to have an understanding of temporal context.  
\end{itemize}

\begin{table}[ht!]
    \centering
    \small
    \begin{tabular}{p{1.5cm}p{5cm}}
    \toprule
    \textbf{Type} & \textbf{Example} \\
    \midrule
    Inferred &  \colorbox{lightgray}{Gen:} Alessandro Del Piero is a former Italian professional football player and a forward. \colorbox{magenta}{Wiki:} Maldini, who was the head coach of the national team, appointed Del Piero as the main forward.  \\
    \midrule
    Subjective & \colorbox{lightgray}{Gen:} Frida Kahlo is widely regarded as \colorbox{red}{the most influential painter} of the 20th century. \\
    \midrule
    Nuanced Difference & \colorbox{lightgray}{Gen:} Louis Pasteur is known as the "Father of Modern Microbiology". \colorbox{magenta}{Wiki:} ... has been honored as the "father of microbiology"... \\
    \midrule
    Temporal Information & \colorbox{lightgray}{Gen:} Michelle Bachelet is currently the President of Chile. \colorbox{magenta}{Wiki:} She served as President of Chile from 2006 to 2010 and from 2014 to 2018... \\
    \bottomrule
    \end{tabular}
    \caption{Categories of special case in annotation.}
    \label{tab:challenge}
\end{table}

Each category presents unique difficulties in determining the factuality of generated content with its evidence source. 

\begin{itemize}[leftmargin=*]
\item \textit{Inferred} connection between generated content and evidence is one of the biggest challenges for both annotators and NLI models, since they need to infer the relationship or the factual basis that links them. This requires a deep understanding of context and the ability to draw inferences from potentially sparse or indirect evidence. 

\item \textit{Subjective} content in generations poses a significant challenge because it introduces personal opinions, emotions, or interpretations that are inherently difficult to verify against factual evidence. For human annotators, this can lead to variability in judgments based on personal biases or interpretations. For NLI models, which are primarily designed for fact-based analysis, handling subjective content requires advanced understanding of sentiment, opinion, and cultural context, areas where current models may fall short.

\item \textit{Nuanced difference} between evidence and generated text highlight the limitations of surface-level text classification approaches in NLI models. Detecting nuanced differences demands a granular analysis of semantics, requiring models to understand context, synonyms, and slight variations in meaning. This challenge underscores the need for more sophisticated NLI models capable of deep semantic analysis and the importance of training annotators to pay attention to detail and understand the significance of minor discrepancies. 

\item \textit{Time-sensitive information} introduces complexity because it requires both annotators and models to have an understanding of temporal context and the ability to evaluate statements within the correct time frame. This can be particularly challenging when information changes over time, requiring up-to-date knowledge and the ability to discern the relevance of temporal qualifiers in text. For NLI models, this underscores the need for dynamic knowledge bases and the ability to reason about time, which are areas where current models may lack proficiency.
\end{itemize}
Overall, these challenges highlight the complexities involved in hallucination detection and the need for advanced capabilities in human annotators and NLI models. 


\section{Metric Calculation Details}
ROUGE scores are calculated with TorchMetrics\footnote{\url{https://github.com/Lightning-AI/torchmetrics}}, and we remove all stopwords before calculating ROUGE-1. Entities are extracted with Spacy's named entity recognizer\footnote{\url{https://spacy.io/api/entityrecognizer}}; we note that this tagger only covers 13 of the 19 languages considered in our experiments. For the NLI-based metric, we finetune the XLMR-large model \cite{conneau2020unsupervised} on the subset of the XNLI dataset \cite{conneau2018xnli} that intersects with the languages used in our experiments. The finetuned model has an average validation accuracy of 85.4\% for the nine intersecting languages.

\section{Additional Results}
We show the pairwise metric results in Table~\ref{tab:pairwise}. We observe similar trends as the reference-based metrics. Also, the average statistics of annotation result are shown in Table~\ref{tab:annot_stat}.

\begin{table}[ht!]
\small
\centering
\begin{tabular}{lcc}
\toprule
\textbf{Metric} & \textbf{Sent-Level} & \textbf{Atomic-Level} \\
\midrule
\# Examples        & 111    & 102   \\
\# Words         & 46.21  & 10.21 \\
\# Evidence      & 2.17   & 1.00  \\
\# Total Facts    & 4.76   & 1.00  \\
Support Rate & 0.35   & 0.29  \\
Contradictory Rate   & 0.15   & 0.24  \\
Unverified Rate    & 0.50   & 0.47  \\
Instruction-conflict Rate    & 0.03   & 0.07  \\
Context-conflict Rate   & 0.13   & 0.06  \\
\bottomrule
\end{tabular}
\caption{Average statistics of Chinese and English annotation data. }
\label{tab:annot_stat}
\end{table}

\begin{table*}[ht!]
    \small
    \centering
    \caption{Results of different pairwise consistency metrics for the BLOOMZ-mt model. "-" indicates they have no coverage for the NER tool we use. All of the ROUGE and Named Entity Overlap results are in percentage (\%).}
    \begin{tabular}{lccccccccccc}
    \toprule
    Language & R1\_F1 & R1\_P & R1\_R & RL\_F1 & RL\_P & RL\_R & NEO\_F1 & NEO\_P & NEO\_R & $\mathrm{DIFF}$ & $\mathrm{UNV}$ \\
    \midrule
    English    & 12.03 & 14.21 & 14.71 & 5.91 & 7.10 & 7.37 & 4.27 & 53.41 & 2.26 & -0.25 & 0.57 \\
    Chinese    & 7.57 & 8.56 & 8.55 & 4.00 & 4.55 & 4.55 & 4.69 & 35.28 & 2.79 & -0.29 & 0.57 \\
    Spanish    & 12.49 & 14.45 & 15.05 & 6.21 & 7.31 & 7.68 & 3.28 & 48.48 & 1.76 & -0.22 & 0.52 \\
    German     & 4.42 & 10.47 & 4.48 & 1.98 & 5.51 & 2.04 & 0.83 & 36.06 & 0.42 & -0.33 & 0.56 \\
    Russian    & 0.15 & 0.19 & 0.13 & 0.00 & 0.00 & 0.00 & 0.48 & 11.28 & 0.25 & -0.09 & 0.55 \\
    Indonesian & 4.74 & 6.23 & 6.52 & 1.59 & 2.13 & 2.27 & -    & -     & -    & -0.16 & 0.68 \\
    Vietnamese & 11.60 & 14.36 & 15.91 & 6.26 & 7.91 & 8.91 & -    & -     & -    & -0.32 & 0.56 \\
    Persian    & 0.00 & 0.00 & 0.00 & 0.00 & 0.00 & 0.00 & -    & -     & -    & -0.31 & 0.52 \\
    Ukrainian  & 0.00 & 0.00 & 0.00 & 0.00 & 0.00 & 0.00 & 0.70 & 20.64 & 0.36 & 0.14  & 0.42 \\
    Swedish    & 10.04 & 14.69 & 10.82 & 8.53 & 13.23 & 9.74 & 1.28 & 45.24 & 0.66 & -0.20 & 0.54 \\
    Thai       & 0.01 & 0.02 & 0.01 & 0.00 & 0.00 & 0.00 & 0.00 & 0.00  & 0.00 & -0.19 & 0.68 \\
    Japanese   & 0.47 & 0.75 & 0.58 & 0.08 & 0.13 & 0.09 & 0.47 & 15.52 & 0.25 & -0.07 & 0.56 \\
    Romanian   & 12.21 & 13.95 & 12.68 & 8.57 & 9.40 & 8.70 & 0.39 & 17.47 & 0.20 & -0.30 & 0.44 \\
    Hungarian  & 0.18 & 0.25 & 0.46 & 0.00 & 0.00 & 0.00 & -    & -     & -    & -0.18 & 0.58 \\
    Bulgarian  & 0.30 & 0.44 & 0.28 & 0.05 & 0.08 & 0.06 & -    & -     & -    & -0.02 & 0.96 \\
    French     & 12.02 & 14.03 & 14.61 & 6.26 & 7.43 & 7.76 & 4.35 & 57.41 & 2.31 & -0.05 & 0.92 \\
    Finnish    & 0.46 & 0.47 & 0.61 & 0.17 & 0.19 & 0.19 & 0.58 & 23.71 & 0.30 & -0.04 & 0.94 \\
    Korean     & 0.19 & 0.26 & 0.17 & 0.00 & 0.00 & 0.00 & 0.24 & 8.48  & 0.12 & 0.25  & 0.63 \\
    Italian    & 4.79 & 7.85 & 5.08 & 2.71 & 4.15 & 2.77 & 1.00 & 30.26 & 0.52 & -0.01 & 0.97 \\
    \bottomrule
    \end{tabular}
    \label{tab:pairwise}
\end{table*}

\begin{table}[ht]
    \small
    \centering
    \begin{NiceTabular}{lccc}
        \toprule
        Metric  &  Pearson & AUC$_\textit{V}$ & AUC$_\textit{UNV}$ \\ 
        \midrule

        Random$_\textit{Sent}$ & -  & 38.74 & 60.05 \\
        $\mathrm{UNV}$$_\textit{Sent}$ & \textbf{0.27} &  32.70 & \textbf{70.31}\\
        $\mathrm{mFACT}$$_\textit{Sent}$& -0.18 &\textbf{45.27}&63.94 \\
        $\mathrm{seahorse}$$_\textit{Sent}$&\colorbox{gray}{-0.03} &41.74 &62.73 \\
        \hdashline
        
        Random$_\textit{Atom}$& - & 68.31	&12.05\\
        $\mathrm{UNV}$$_\textit{Atom}$& 0.09 &73.49 &20.19 \\
        $\mathrm{mFACT}$$_\textit{Atom}$& \textbf{-0.17} &\textbf{79.03} &\textbf{32.01}  \\
        $\mathrm{seahorse}$$_\textit{Atom}$& -0.12 &78.24 & 30.93  \\

        \comment{
        \midrule
        \multicolumn{5}{c}{\textit{At least 50\% unverifiable facts} ($N_{nv}/N_t \geq 50\%$)}\\
        Random$_\textit{Sent}$ & - & 47.47 &44.09  \\
        $\mathrm{UNV}$$_\textit{Sent}$ & - & 55.63 &\textbf{54.63} \\
        $\mathrm{mFACT}$$_\textit{Sent}$& -&\textbf{80.93} &44.51 \\
        $\mathrm{seahorse}$$_\textit{Sent}$& - &72.97 & 31.27 \\
        \midrule
        \multicolumn{5}{c}{\textit{100\% unverifiable facts} ($N_{nv}=N_t$)}\\
        Random$_\textit{Sent}$ & - & 91.42 & 6.87\\
        $\mathrm{UNV}$$_\textit{Sent}$ & - & 97.31 & 3.80 \\
        $\mathrm{mFACT}$$_\textit{Sent}$& - &96.00 &6.57 \\
        $\mathrm{seahorse}$$_\textit{Sent}$& - &93.99 &3.86\\
        }
        \bottomrule
    \end{NiceTabular}
    \caption{Results of unverifiable hallucination detection in human evaluation in \underline{English}, with the best agreement with humans indicated in bold. $\textit{Sent}$ denotes sentence-level detection, and $\textit{Atom}$ denotes atomic-fact-level detection. $\textit{UNV}$ denotes unverifiable hallucination, and $\textit{V}$ denotes verifiable. The number in \colorbox{gray}{gray} have the \textit{p-values} larger than 0.05.} 
    \label{tab:unverify_en}
\end{table}

\begin{table}[ht]
    \small
    \centering
    \begin{NiceTabular}{lccc}
        \toprule
        Metric  &  Pearson & AUC$_\textit{V}$ & AUC$_\textit{UNV}$ \\ 
        \midrule

        Random$_\textit{Sent}$ & -  &  39.24 & 50.22 \\
        $\mathrm{UNV}$$_\textit{Sent}$ & 0.31 & 30.27 &\textbf{76.12} \\
        $\mathrm{mFACT}$$_\textit{Sent}$& \textbf{-0.32} &\textbf{56.54} &60.42 \\
        $\mathrm{seahorse}$$_\textit{Sent}$& \colorbox{gray}{-0.11} &41.45 &63.54\\
        \hdashline
        Random$_\textit{Atom}$& - & 71.93	&17.88\\
        $\mathrm{UNV}$$_\textit{Atom}$& 0.12 &79.26 &27.36 \\
        $\mathrm{mFACT}$$_\textit{Atom}$& \textbf{-0.19} &\textbf{86.77} &\textbf{34.38}  \\
        $\mathrm{seahorse}$$_\textit{Atom}$& -0.14 &80.76 & 30.92  \\

        \comment{
        \midrule
        \multicolumn{5}{c}{\textit{At least 50\% unverifiable facts} ($N_{nv}/N_t \geq 50\%$)}\\
        Random$_\textit{Sent}$ & - & 47.47 &44.09  \\
        $\mathrm{UNV}$$_\textit{Sent}$ & - & 55.63 &\textbf{54.63} \\
        $\mathrm{mFACT}$$_\textit{Sent}$& -&\textbf{80.93} &44.51 \\
        $\mathrm{seahorse}$$_\textit{Sent}$& - &72.97 & 31.27 \\
        \midrule
        \multicolumn{5}{c}{\textit{100\% unverifiable facts} ($N_{nv}=N_t$)}\\
        Random$_\textit{Sent}$ & - & 91.42 & 6.87\\
        $\mathrm{UNV}$$_\textit{Sent}$ & - & 97.31 & 3.80 \\
        $\mathrm{mFACT}$$_\textit{Sent}$& - &96.00 &6.57 \\
        $\mathrm{seahorse}$$_\textit{Sent}$& - &93.99 &3.86\\
        }
        \bottomrule
    \end{NiceTabular}
    \caption{Results of unverifiable hallucination detection in human evaluation in \underline{Chinese}, with the best agreement with humans indicated in bold. All the \textit{p-values} of the correlation is less than 0.05. $\textit{Sent}$ denotes sentence-level detection, and $\textit{Atom}$ denotes atomic-fact-level detection. $\textit{UNV}$ denotes unverifiable hallucination, and $\textit{V}$ denotes verifiable. The number in \colorbox{gray}{gray} have the \textit{p-values} larger than 0.05.} 
    \label{tab:unverify_zh}
\end{table}

\section{Generation Prompt Templates}
\begin{figure}[ht]
    \centering
    \includegraphics[width=0.95\columnwidth]{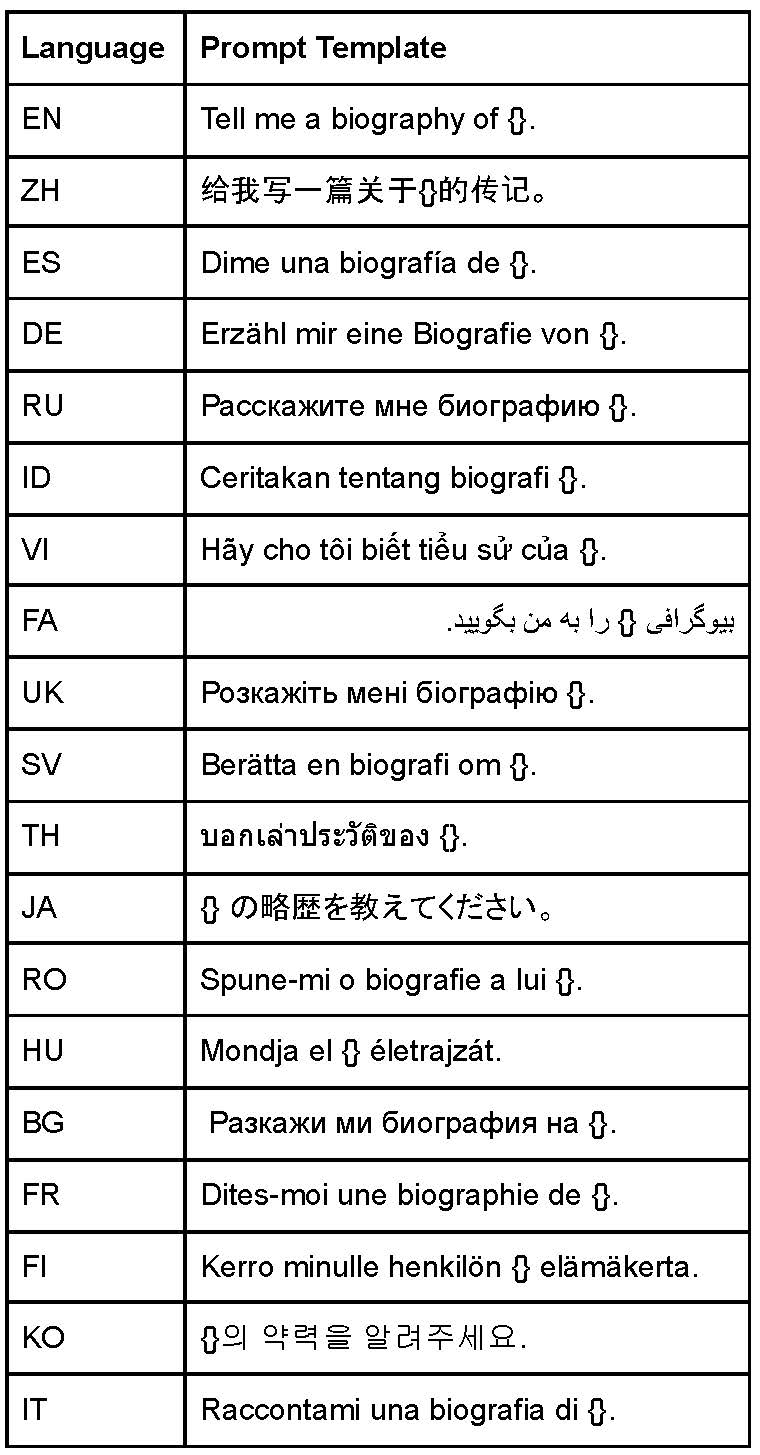}
    \caption{Prompt templates of all languages used in generating biography. \{\} represents human names.}
    \label{fig:biography_prompts}
\end{figure}

We present the full set of prompt templates for all languages from Section 3 in Figure~\ref{fig:biography_prompts}. 

\begin{table*}[ht]
\centering
\small

\begin{NiceTabular}{l|ccc|ccc}

\toprule
     & \multicolumn{3}{c}{Sentence-Level} &  \multicolumn{3}{c}{Atomic-Fact-Level} \\
    Metric  &  Pearson & AUC$_{\textit{F}}$ & AUC$_{\textit{NF}}$  & Pearson & AUC$_{\textit{F}}$ & AUC$_{\textit{NF}}$   \\ 
\comment{
\midrule
Random      & -           & 10.84   & 82.86  & - &52.11&43.56\\ 
 \hdashline 
\textit{Pairwise} \\
R1-P. &  0.08$^\dagger$\       & 19.78   & 81.48  & 0.10  & 51.23 & 44.23   \\ 
RL-P. &  0.11$^\dagger$\        & 20.04   & 83.10 & 0.12$^\dagger$\ & 52.18 & 42.83 \\ 
NEO-P. &  0.14$^\dagger$\      & 17.49   & 80.84    & 0.09 & 53.09 & 45.32\\ 
$\mathrm{DIFF}$ & 0.21 & 38.46  & 89.49   & 0.19 &57.46 &54.41 \\ 
$\mathrm{ENT}$ & 0.31 & 40.32 & 90.86 & 0.23 & 60.71 & 57.48\\
$\mathrm{CON}$ & 0.11  & 16.47   & 80.41 & -0.01& 51.49 & 52.16 \\ 
}
\midrule
\textit{Reference} \\

random & -    & 18.34   & 82.56     & - & 54.28 & 48.31 \\ 
$\mathrm{DIFF}$    & 0.28        & \textbf{49.11}  & \textbf{92.41} & 0.29 &63.93 &62.49 \\ 
$\mathrm{ENT}$ & \textbf{0.31} & 45.43 & 90.27  & \textbf{0.33} &\textbf{66.67} & \textbf{63.69}  \\
$\mathrm{CON}$ & -0.14  & 31.56   & 92.20   &-0.21& 60.18 & 56.90  \\ 
$\mathrm{mFact}$ & 0.31  & 31.70  & 85.52   & 0.28& 64.84& 60.10\\
$\mathrm{Seahorse}$ & 0.08$^\dagger$\  & 29.28   & 85.80   & 0.09$^\dagger$\ & 60.94 & 57.44 \\
\bottomrule
\end{NiceTabular}
\caption{Sentence- and atomic-fact-level verifiable hallucination detection in human evaluation in \underline{English} comparing automatic metrics and human support rate, with the best agreement with humans indicated in bold. \textit{F} denotes factual examples and \textit{NF} denotes non-factual examples. $^\dagger$\textit{p-values} of this correlation is larger than 0.05.} 
\label{tab:veri-detection-en}
\end{table*}

\end{document}